\pdfoutput=1

\documentclass{article}

\usepackage{arxiv}

\usepackage{multicol}
\usepackage[bookmarks=true]{hyperref}

\usepackage{graphicx}
\usepackage{amsfonts}         
\usepackage{amsmath} 
\usepackage{amssymb} 
\usepackage{amsthm}
\usepackage{diffcoeff}
\usepackage{subfigure}
\usepackage{tabularx}
\usepackage[symbol]{footmisc}
\usepackage{mathtools}
\usepackage{dsfont}
\usepackage{bigints}
\usepackage{multirow}
\usepackage{epstopdf}

\DeclareMathOperator{\vecop}{vec}

\title{Safe Optimal Control under Parametric Uncertainties}

\author{Hemanth Sarabu\thanks{Equal contributions (listed alphabetically)}\\
Symbio Robotics\\
\texttt{hemanth@symb.io}\\
  \And
  Venkata Ramana Makkapati\footnotemark[1]\\
  Georgia Tech\\
  \texttt{mvramana@gatech.edu}\\
  \And
  Vinodhini Comandur\footnotemark[1]\\
  Georgia Tech\\
  \texttt{vinodhini@gatech.edu}\\
  \And
  Panagiotis Tsiotras\\
  Georgia Tech\\
  \texttt{tsiotras@gatech.edu}
  \And
  Seth Hutchinson\\
  Georgia Tech\\
  \texttt{seth@gatech.edu}
}

\begin{document}

\maketitle

\begin{abstract}
We address the issue of safe optimal path planning under parametric uncertainties using a novel regularizer that allows trading off optimality with safety.
The proposed regularizer leverages the notion that collisions may be modeled as constraint violations in an optimal control setting in order to produce open-loop trajectories with reduced risk of collisions. 
The risk of constraint violation is evaluated using a state-dependent \emph{relevance function} and first-order variations in the constraint function with respect to parametric variations. 
The approach is generic and can be adapted to any optimal control formulation that deals with constraints under parametric uncertainty. 
Simulations using a holonomic robot avoiding multiple dynamic obstacles with uncertain velocities are used to demonstrate the effectiveness of the proposed approach. 
Finally, we introduce the \emph{car vs. train problem} to emphasize the dependence of the resultant risk aversion behavior on the form of the constraint function used to derive the regularizer.  
\end{abstract}


\section{Introduction}
The tension between optimality and safety is often evident in robotics---particularly for applications that have stringent performance requirements---under conditions for which uncertainties in sensing, environment models, and control effectiveness are unavoidable \cite{zhang2012improving, hauser2012responsiveness, lengagne2007, luders2013robust}. 
For all but the simplest applications, optimal solutions tend to bring the robot dangerously close to the operational safety margins. 
For example, it is well known that the shortest path for a mobile robot in a polygonal environment lies in the visibility graph which implies that the optimal path would contact the obstacles while traversing the path \cite{mitchell2000geometric}. 
While in practice it is typical to perturb paths slightly such that they do not reach the constraint boundaries, this safety strategy raises a number of significant questions:
How should one perform these perturbations?
How should one balance the cost of violating constraints
against reduced performance?
And, perhaps most importantly, how can one provide a principled
evaluation of the effects of uncertainty with respect to the
trade-offs between optimality and safety, and adjust the path to optimally balance between the two?
It is this latter question that we address in the present paper.

Our approach exploits recent results in the area of desensitized optimal control (DOC) \cite{makkapati2019c,ramana2018doc}.
DOC techniques modify the nominal optimal trajectory such that it is \emph{less sensitive} with respect to uncertain parameters.
This involves constructing an appropriate sensitivity cost which, when penalized, provides solutions that are relatively insensitive to parametric variations.
The process of constructing desensitized trajectories can be also understood as a way to impose smoothness through the regularization of the sensitivities \cite{ohlsson2010regularization}.
DOC methods take a global view of parametric uncertainty, attempting to provide robust solutions along the entire trajectory \cite{ramana2018doc,shen2010desensitizing}.
In contrast, in this paper we consider explicitly the effects of parametric uncertainty for those portions of the trajectory that approach (or touch) constraint boundaries, essentially ignoring parametric uncertainties for states that are inherently safe even under large parametric variations.
In this way, we focus both computation and control effort only on those
areas that are most crucial for overall system safety. 

Probabilistic methods are a popular choice to address planning in uncertain dynamic environments \cite{thrun2000probabilistic}.
In the past, Gaussian processes have been employed to model uncertainty and to obtain safe trajectories \cite{aoude2013probabilistically, xu2014motion, van2011lqg, akametalu2014reachability}.
Techniques involving POMDPs \cite{luo2018porca, zhu1991hidden}, occupancy grids \cite{fulgenzi2007dynamic}, intent-based threat estimation \cite{aoude2010threat}, replanning \cite{berg2006anytime, zucker2007multi}, and feedback coupled with estimation \cite{toit2010robotic} are many variants in the class of probabilistic methods.
Alternatively, reachability analysis \cite{fridovich2019safely}, artificial potential fields (APFs) \cite{mora2008path}, and barrier functions \cite{rafieisakhaei2016non} have also been utilized.
Our approach fundamentally differs from the above techniques in its formulation; while prior work widely employed probabilistic techniques, the proposed formulation in this paper is deterministic.
In order to address planning under limited sensing and feedback capabilities, we provide safe open-loop trajectories that have guarantees on optimality by treating the uncertainty to be parametric in nature, and by examining sensitivities with respect to parameter variations.

Formally, we consider uncertainties to be parametric in nature, where the nominal value of the uncertain parameter is available.
Using sensitivity functions \cite{khalil}, we first capture the variations in the constraint function under parametric variations, defined as \emph{constraint sensitivity}.
The variations are then weighted using a relevance function to obtain the \emph{relevant constraint sensitivity} (RCS), and construct a regularizer that captures the risk of constraint violation. 
The characteristics of the regularizer are discussed by analyzing its performance in simple path planning problems.
Finally, we evaluate the proposed technique on path planning problems in environments containing up to ten dynamic obstacles having uncertain velocities.

The rest of the paper is organized as follows. 
Section \ref{sec:prelims} introduces sensitivity functions and the framework of DOC.
Section \ref{sec:main_idea} presents the main idea of the paper, involving the construction of an appropriate regularizer that provides open-loop trajectories with lower chance of constraint violation under parametric uncertainties.
In Section \ref{sec:analysis}, we first analyze the proposed approach by applying it on simple path planning problems with one dynamic obstacle, and then present the results obtained from experiments on environments with multiple uncertain dynamic obstacles.
Section \ref{sec:conclusion} concludes the paper.


\section{Preliminaries}
\label{sec:prelims}

\subsection{Standard Optimal Control Framework}

Consider the standard optimal control problem of minimizing the cost
\begin{align}
\mathcal{J}(u) = \phi(x(t_f),t_f) + \int_{t_0}^{t_f}L(x(t),u(t),t) \, \text{d}t,
\label{eq:cost}
\end{align}
subject to
\begin{align}
&\dot{x}(t) = f(x(t),p,u(t),t), \quad x(t_0) = x_0, \label{eq:ini-dynamics} \\
&g(x(t),p,t) \leq 0, \label{eq:state_constr}\\
&\psi(x(t_f),t_f) = 0, \label{eq:ter_cond}
\end{align}
where $t \in [t_0,\,t_f]$ denotes time, with $t_0$ being the fixed initial time and $t_f$ being the final time, $x(t) \in \mathbb{R}^n$ denotes the state, with $x_0$ being the fixed state at $t_0$, and $p \in \mathcal{P} \subset \mathbb{R}^\ell$ are model parameters. 
The control $u \in \mathcal{U}=\left\lbrace u:[t_0,t_f] \rightarrow U \text{ is Piecewise Continuous},~u(t) \in  U,\right.$ $\left.t \in \left[t_0,t_f\right]\right\rbrace$
with $U \subseteq \mathbb{R}^{m}$, the set of allowable values of $u(t)$,
$\phi:\mathbb{R}^{n}\times [t_0,t_f] \rightarrow \mathbb{R}$ is the terminal cost function, and $L:\mathbb{R}^{n}\times\mathbb{R}^{m}\times [t_0,t_f] \rightarrow \mathbb{R}$ is the running cost.
Here $g: \mathbb{R}^n \times \mathbb{R}^\ell \times [t_0,t_f] \rightarrow \mathbb{R}^k$ is a function denoting $k$ state inequality constraints.
Finally, $\psi(x(t_f),t_f) = 0$ denotes the terminal condition at time $t = t_f$.

The aforementioned optimal control problem (\ref{eq:cost})-(\ref{eq:ter_cond}) is to be solved by finding the optimal control $u^* \in \mathcal{U}$ that minimizes the cost function in (\ref{eq:cost}), given the constraints (\ref{eq:ini-dynamics})-(\ref{eq:ter_cond}). 
The solution defines the optimal state trajectory $x^*(t)$, $t \in [t_0,t_f]$, satisfying $\dot{x}^*(t) = f(x^*(t), p, u^*(t), t)$ subject to $x^*(t_0) = x_0.$
The system dynamics $f(x,p,u,t)$ contains the model parameters, $p \in \mathcal{P}$, which are assumed to be constant.
In general, the optimal solution $(x^*(t),u^*(t))$ is sensitive to modeling errors and, if changes in the parameters $p$ occur at any time $t \in [t_0,t_f]$, satisfaction of the constraints (\ref{eq:state_constr}) or (\ref{eq:ter_cond}) is not guaranteed.

In the case of path planning with dynamic obstacles collision avoidance is of paramount importance.
Obtaining safe trajectories under parametric uncertainties in the obstacles' motion is therefore a necessity.
In the optimal control framework discussed above, the constraint function in (\ref{eq:state_constr}) can be used to enforce collision avoidance for the path planning problem.
Consequently, penalizing a risk measure that captures the possibility of constraint violation under parametric variations may provide the desired safe (e.g., lower chance of constraint violation) trajectories.

With the motivation to minimize the dispersion in the optimal trajectory under parameter uncertainties, an augmented cost function using sensitivity functions has been constructed in \cite{ramana2018doc}.
In that paper, sensitivity functions were used to impose the desired risk measure for constraint violation.
To this end, we first discuss the approach in \cite{ramana2018doc} along with the theory behind sensitivity functions.

\subsection{Sensitivity Functions and DOC}

Consider the dynamics in (\ref{eq:ini-dynamics}), and assume variations in the model parameters $p \in \mathcal{P}$, with $p = p_0$ being the nominal value of the parameter vector.
Furthermore, assume that $f(x,p,u,t)$ is continuous in $(x,p,u,t)$, and continuously differentiable with respect to $x$ and $p$ for all $(x,p,u,t) \in \mathbb{R}^n \times \mathcal{P} \times U \times [t_0,t_f]$.
The solution to the differential equation from the initial condition $x_0$ with control input $u \in \mathcal{U}$ is given by
\begin{align}
x(p,t) = x_0 + \int_{t_0}^{t} f(x(p,\tau),p,u(\tau),\tau) \, \text{d}\tau.
\end{align}
Since $f(x,p,u,t)$ is differentiable with respect to $p$, it follows that
\begin{align}
\frac{\partial x(p,t)}{\partial p} &= \int_{t_0}^{t} \left[ \frac{\partial f(x(p,\tau),p,u(\tau),\tau)}{\partial x} \frac{\partial x(p,\tau)}{\partial p} + \frac{\partial f(x(p,\tau),p,u(\tau),\tau)}{\partial p} \right]  \, \text{d}\tau.
\end{align}
Taking the derivative with respect to $t$, we obtain
\begin{align}
\frac{\mathrm{d}}{\mathrm{d} t} \left[ \frac{\partial x(p,t)}{\partial p}\right]  &= \frac{\partial f(x,p,u(t),t)}{\partial x} \frac{\partial x(p,t)}{\partial p} + \frac{\partial f(x,p,u(t),t)}{\partial p}.  \label{eq:sensi_partial} 
\end{align}
Evaluating (\ref{eq:sensi_partial}) at the nominal conditions ($p = p_0$), the dynamics for the \textit{parameter sensitivity function} $S:[0,\infty)\rightarrow\mathbb{R}^{n\times \ell}$
\begin{align}
S(t) = \dfrac{\partial x (p,t)}{\partial p}\bigg\rvert_{x = x(p_0,t)} \label{eq:SF}
\end{align}
can be obtained as
\begin{align}
\dot{S}(t) = A(t)S(t) + B(t), \quad S(t_0) = 0_{n \times \ell}, \label{eq:sensi_eq}
\end{align}
where
\begin{align}
A(t) = \dfrac{\partial f(x,p,u(t),t)}{\partial x}\bigg\rvert_{x = x(p_0,t),\, p = p_0}, \quad
B(t) = \dfrac{\partial f(x,p,u(t),t)}{\partial p}\bigg\rvert_{x = x(p_0,t),\, p = p_0}.
\end{align}

Since the initial state is given (fixed), the initial condition for the sensitivity function is the zero matrix, and (\ref{eq:sensi_eq}) is called the \emph{sensitivity equation} in the literature \cite{khalil}.
To compute the sensitivity function over time, the state $x$ has to be propagated using the dynamics in (\ref{eq:ini-dynamics}) under nominal conditions,
\begin{align}
\dot{x} = f(x,p_0,u,t), \quad x(t_0) = x_0. \label{eq:x_dynamics}
\end{align}
From the properties of continuous dependence with respect to the parameters and the differentiability of solutions of ordinary differential equations, and for sufficiently small variations in $p_0$, the solution $x(p,t)$ can be approximated by
\begin{align}
x(p,t) \approx x(p_0,t) + S(t)(p - p_0). \label{eq:sensi_approx}
\end{align}
This is a first-order approximation of $x(p,t)$ about the nominal solution $x(p_0,t)$.

For the optimal control problem (\ref{eq:cost})-(\ref{eq:ter_cond}) in the previous subsection, an approach to construct a \emph{desensitized optimal control} (DOC) scheme that reduces the dispersion of the optimal trajectories under parametric uncertainties is to minimize the augmented cost function
\begin{align}
\mathcal{J}_s(u) = \mathcal{J}(u) +\int_{t_0}^{t_f}  \|\vecop S(t)\|_{Q}^{2} \, \text{d}t,  \label{eq:TD_aug}
\end{align}
with an augmented state $[x^\top~(\vecop S)^\top]^\top$, whose dynamics are obtained from (\ref{eq:x_dynamics}), (\ref{eq:sensi_eq}), and the constraints (\ref{eq:state_constr}), (\ref{eq:ter_cond}).
Here, $\vecop S \in \mathbb{R}^{n\ell}$ denotes vectorization of the matrix $S$.
Equation (\ref{eq:TD_aug}) minimizes the original cost function in (\ref{eq:cost}), while penalizing the sensitivity of the state with respect to the parameters along the optimal trajectory. The weighting factor for the sensitivity cost, $Q \geq 0$, can be tuned to balance between minimizing the original cost and minimizing the sensitivity cost. 
In the next section, we develop a scheme to generate constraint desensitized trajectories by penalizing a risk measure that is defined using sensitivity functions.


\section{Constraint Desensitized Path Planning}
\label{sec:main_idea}

For the optimal control problem (\ref{eq:cost})-(\ref{eq:ter_cond}), assuming the constraint function $g(x,p,t)$ is a smooth function in $x$, a na\"ive approach to obtain conservative trajectories to address constraint violation under parametric uncertainties would be to penalize the \emph{constraint sensitivity} matrix, defined as
\begin{align}
    S_g(t) &= \diffp{g(x(p,t),p,t)}{p}\bigg\rvert_{p = p_0} \nonumber \\
    &= \left(\diffp{g(x,p,t)}{x}S(t) + \diffp{g(x,p,t)}{p}\right)\bigg\rvert_{x = x(p_0,t),p = p_0},
\end{align}
by constructing the augmented cost
\begin{align}
\mathcal{J}_g(u) = \mathcal{J}(u) + \int_{t_0}^{t_f} \|\vecop S_g(t)\|_{Q}^{2} \, \text{d}t,  \label{eq:naive_cost}
\end{align}
where $Q \geq 0$.
By minimizing the cost in (\ref{eq:naive_cost}), one attempts to minimize the variation in the constraint value with respect to variations in the parameter for all times.
However, it is clear that the variation in the constraint value when the system is far from the constraint boundary is not as important as when the system is close to the constraint boundary.
For example, in the path planning problem with a dynamic obstacle, a larger variation in the constraint value may be acceptable when the agent is far from the obstacle, but a collision may result due to even
relatively small variations when the agent is near the constraint boundary. 
Weighting the sensitivity of the constraint value equally in both cases using a running cost function may 
therefore lead to solutions that are highly sensitive near the constraint boundary.

To account for the fact that the constraint variations are more likely to cause constraint violations when the system is closer to the constraint boundary, we introduce a relevance function $\rho:\mathbb{R} \rightarrow [0,\infty)$ of the form
\begin{align}
    \rho(z) = \begin{cases}
    \tilde{\rho}(z), \quad \text{if } z\leq 0,\\
    \tilde{\rho}(0), \quad \text{if } z > 0,
    \end{cases}
\end{align}
where $\tilde{\rho}:\mathbb{R} \rightarrow [0,\infty)$ is a continuous function that is monotonically increasing over the interval $(-\infty,0]$, that is, $\tilde{\rho}(z) \geq \tilde{\rho}(y)$, if $z > y$ for all $z,y \le 0$. 
Examples of $\tilde{\rho}(z)$ include $e^{-z^2}$ (Gaussian), $\text{max}(0,1-|z|)$ (Hat function), $1/(1+z^2)$, etc.

Next, we define the \emph{relevant constraint sensitivity} (RCS) matrix $S_r:[0,\infty) \rightarrow \mathbb{R}^{k\times \ell}$ as
\begin{align}
    S_r(t) = RS_g(t),
\end{align}
where $R = \text{diag}\left(\rho(g_1(x(p,t),p,t)),\dots,\rho(g_k(x(p,t),p,t))\right)$.
Henceforth for the purpose of the analysis, and unless stated otherwise,  
the derivative of the logistic function $s(z) = 1/(1 + e^{-z})$ is chosen as the candidate relevance function, that is,
\begin{align}
    \tilde{\rho}(z) = s(z)(1 - s(z)).
\end{align}
Note that the derivative of the logistic function has a symmetric ``bell-shape'' with the maximum at $z = 0$, and decaying tails. 
The relevance function allows one to penalize sensitivities according to their relevance  with respect to potential constraint violation.
The impact of the choice of the relevance function on constraint desensitization is briefly discussed in Section \ref{sec:analysis}.A. 
The sensitivity matrix $S_r$ captures the idea of giving more importance to variations near the constraint boundary.

Finally, we propose to solve the optimal control problem with the augmented cost function 
\begin{align}
    \mathcal{J}_a(u) = \mathcal{J}(u) + \int_{t_0}^{t_f} \|\vecop S_r(t)\|_{Q}^{2} \, \text{d}t,  \label{eq:CDP_cost}
\end{align}
the dynamics in (\ref{eq:x_dynamics}) and (\ref{eq:sensi_eq}), and the constraints (\ref{eq:state_constr}) and (\ref{eq:ter_cond}) to construct trajectories that address constraint violation under parametric uncertainties.
Hereafter, the term $\int_{t_0}^{t_f} \|\vecop S_r(t)\|_{Q}^{2} \, \text{d}t$ in (\ref{eq:CDP_cost}) will be referred to as the RCS cost.


\section{Numerical Examples}
\label{sec:analysis}

In this section, we apply the proposed approach on simple test examples to analyze the optimal trajectories obtained by penalizing RCS. 
First, we analyze the claim of penalizing RCS over constraint sensitivity using a 2D path planning problem involving a dynamic obstacle with uncertainty in its speed.
Subsequently, the effect of various constraint forms that represent the collision avoidance condition, chosen from a set of valid ones, is studied.
We then stress upon the need to select an appropriate constraint function to construct RCS using the car vs. train problem, and finally, the trade-off studies with multiple obstacles are presented.
The videos demonstrating the optimal trajectories for the example problems discussed in this section can be found on the web (\url{https://youtu.be/zCvuIQSMzlw}).

\subsection{2D Time-Optimal Problem}\label{subsec:2D_PP}

Consider the following 2D time-optimal path planning problem with the agent dynamics and initial conditions
\begin{align}
    \dot{x}_a(t) &= v_a \cos(\theta(t)), \quad x_a(0) = a_0, \label{eq:xa_dyn}\\
    \dot{y}_a(t) &= v_a \sin(\theta(t)), \quad y_a(0) = b_0,\label{eq:ya_dyn}
\end{align}
where $(x_a,y_a)$ denotes the agent's position, $v_a$ is the agent's speed, and $\theta(t) \in [0,2\pi)$ is the agent's heading (control). 
The agent intends to reach $(a_f,b_f)$ in minimum time, while avoiding a dynamic circular obstacle that is moving parallel to the y-axis with a constant speed $v_o$, and dynamics given by
\begin{align}
    \dot{y}_o(t) &= -v_o, \quad y_o(0) = c,\label{eq:yo_dyn}
\end{align}
where $(x_o,y_o)$ denotes the obstacle's position.
With $x = [x_a,y_a,y_o]^\top$ as the state vector, the constraint for collision avoidance can be expressed as
\begin{align}
    g(x) = r_o - \left[(x_a - x_o)^2 + (y_a - y_o)^2\right]^{1/2} \leq 0, \label{eq:2D_constraint}
\end{align}
where $r_o$ is the safe distance.
For this problem, we assume that the obstacle's speed $v_o$ is the uncertain parameter.
Henceforth, the time and parameter dependency of the elements in the state vector are dropped for brevity.
Since the problem has simple dynamics, a closed form expression to the sensitivity of the constraint function $g(x)$ with respect to the uncertain parameter $v_o$ is given by
\begin{align}
    S_g(t) = \diffp{g(x)}{v_o} &= -\frac{(y_a - y_o)t}{\left[(x_a - x_o)^2 + (y_a - y_o)^2\right]^{1/2}}.
\end{align}

\begin{figure}[htb!]
    \centering
    \subfigure[abs($S_g$)]{\includegraphics[width = 0.4\textwidth]{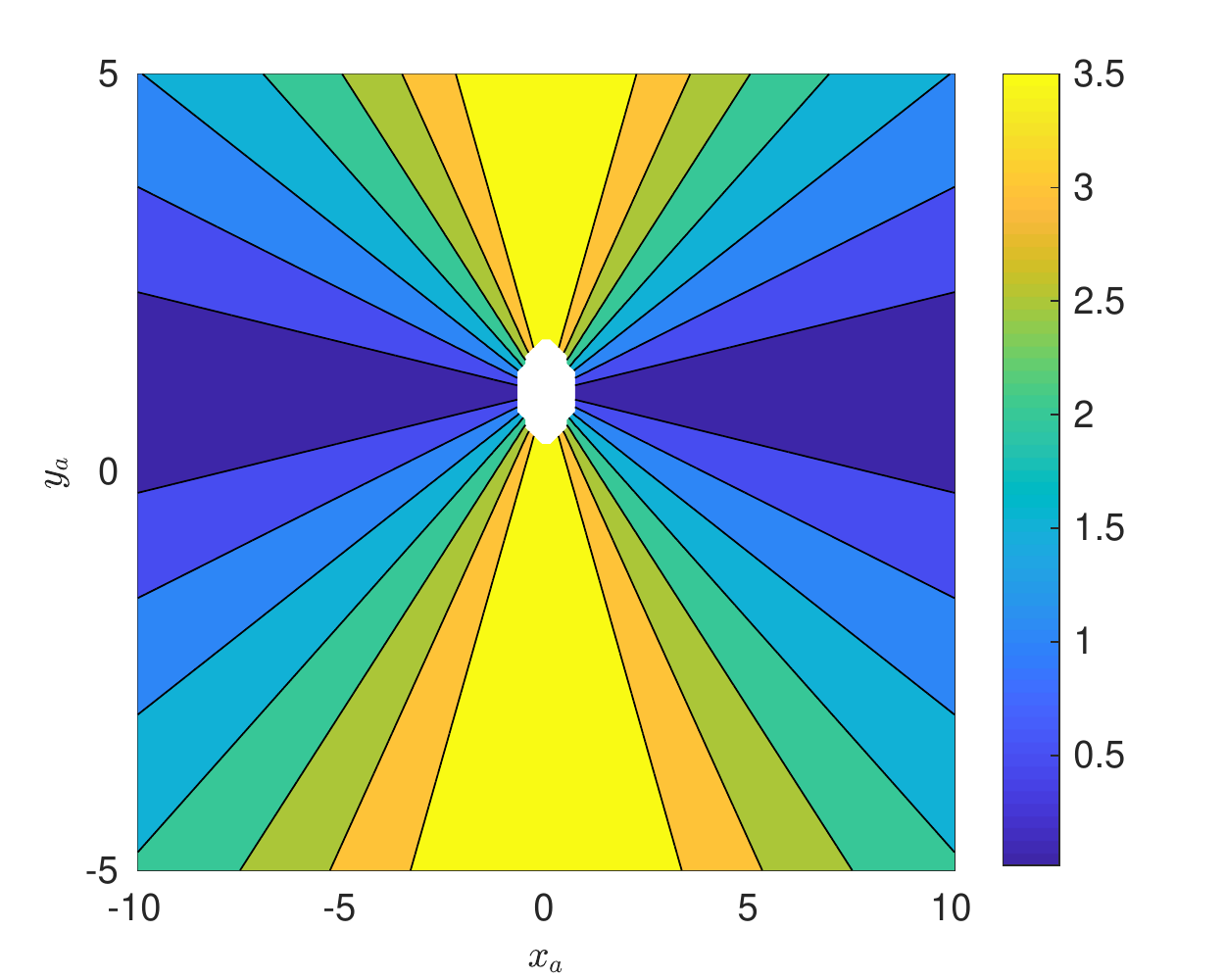}}
    \subfigure[abs($S_\ell$)]{\includegraphics[width = 0.4\textwidth]{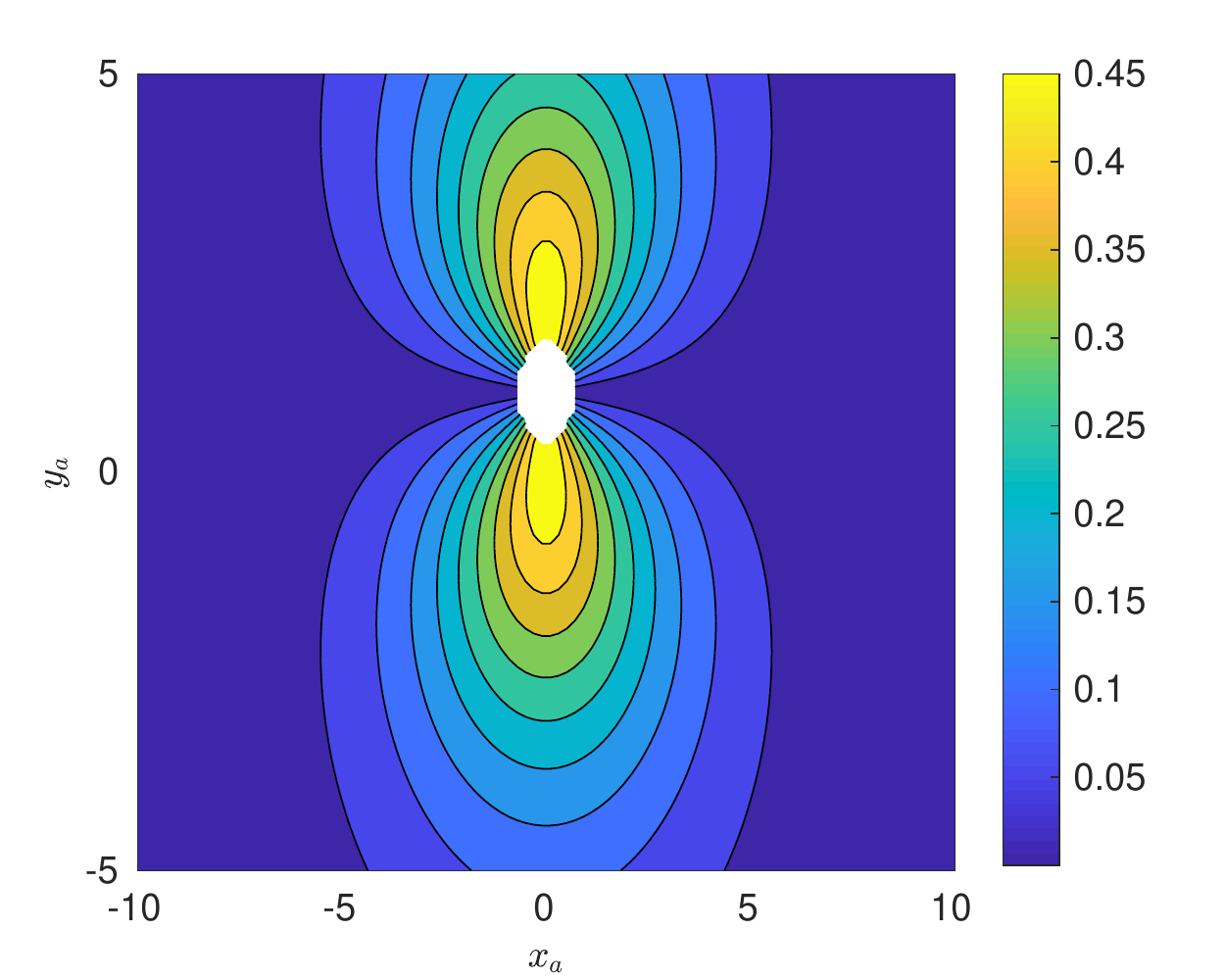}}
    \caption{Absolute values of $S_g$ and $S_\ell$ as a function of the agent's position.}
    \label{fig:2D_sensi}
\end{figure}

The effect of incorporating the relevance function into the proposed scheme is analyzed by comparing the constraint sensitivity and RCS.
In this regard, the obstacle's position is fixed at $(0,1)$ with $r_o = 0.6$, and since the constraint sensitivity varies linearly with time, $t=1$ is chosen.
Figure~\ref{fig:2D_sensi} presents the absolute values of the sensitivities ($S_g$ and $S_r$) over the mesh generated to represent the agent's position.
The white circular area in the middle represents the infeasible region.
It can be observed that RCS ($S_r$) is activated when the agent gets closer to the obstacle, as opposed to the constraint sensitivity, which only captures the sensitivity in the constraint value.
Furthermore, it has also been observed (though not presented here for the sake of brevity) that by just penalizing the constraint sensitivity using the cost in (\ref{eq:naive_cost}), conservative trajectories cannot be obtained as the sensitivity profile over-constrains the problem (see Figure~\ref{fig:2D_sensi}(a)).
In the case of penalizing RCS, the agent has sufficient incentive to move away from the obstacle as the variations closer to the constraint boundary now incur a higher penalty (see Figure~\ref{fig:2D_sensi}(b)).

\begin{figure*}[htb!]
    \centering
    \subfigure[Trajectories]{\includegraphics[width = 0.3\textwidth]{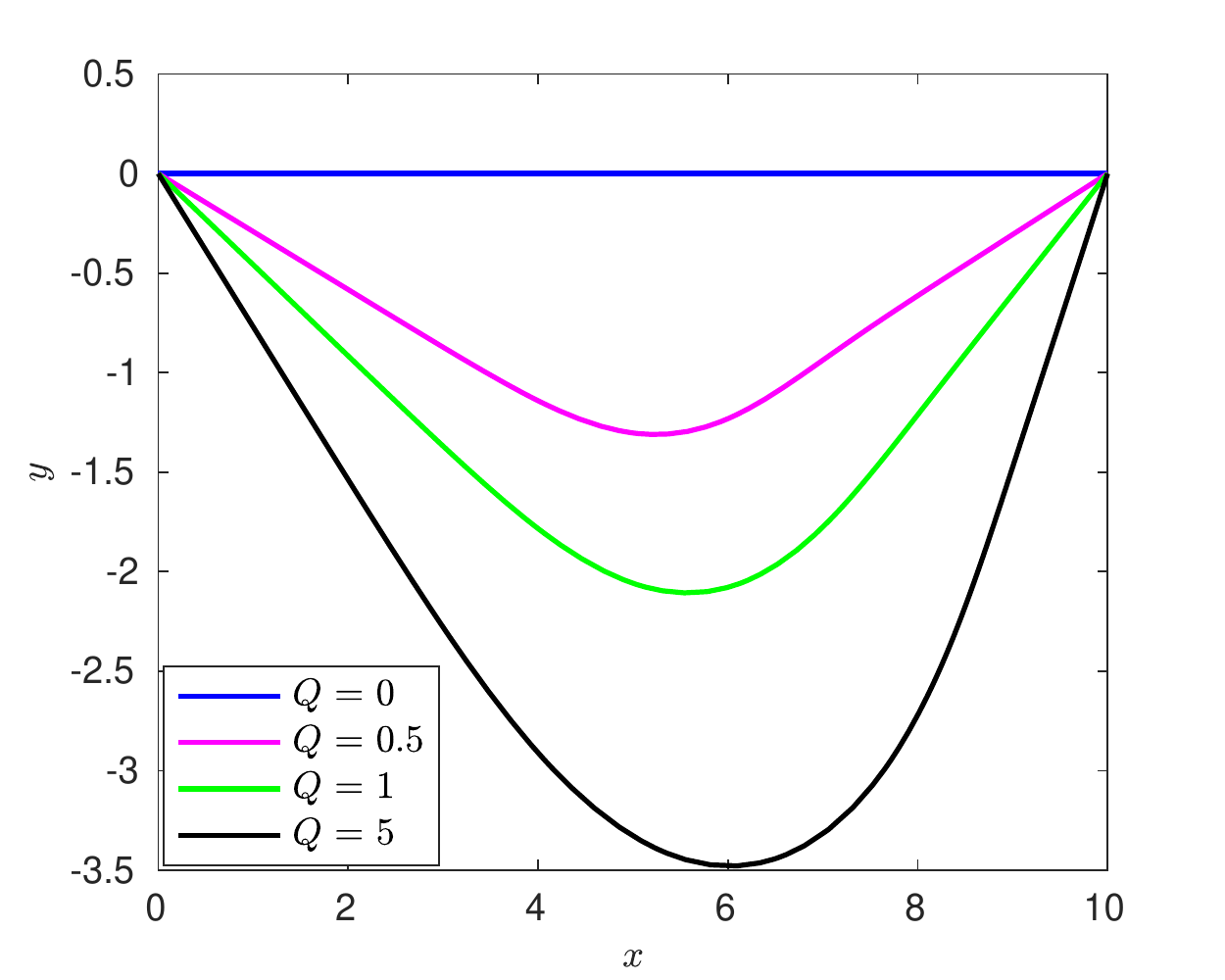}}
    \subfigure[RCS - $S_r^2(t)$]{\includegraphics[width = 0.3\textwidth]{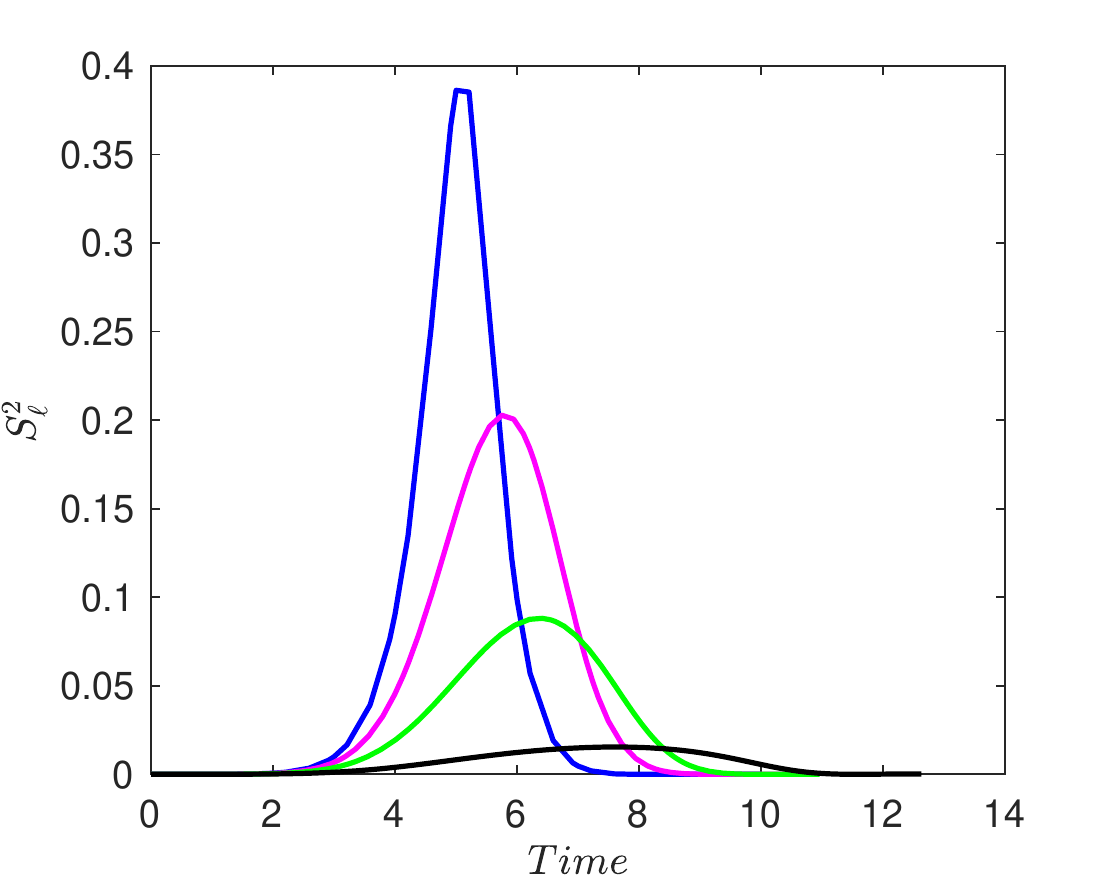}}
    \subfigure[Constraint sensitivity - $S_g^2(t)$]{\includegraphics[width = 0.3\textwidth]{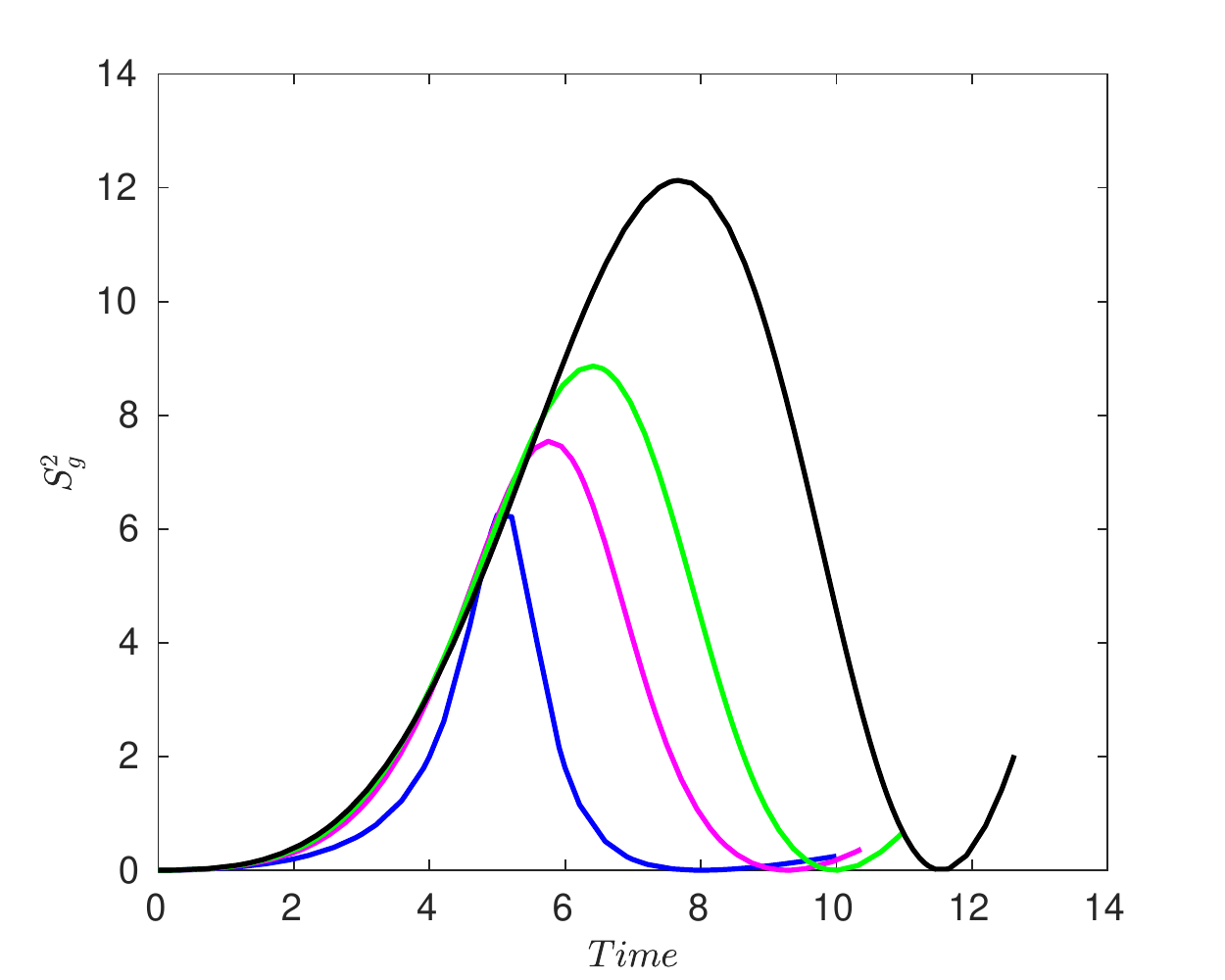}}
    \caption{Results for constraint desensitized 2D path planning}
    \label{fig:2D_oneobs}
\end{figure*}
\begin{figure}[htb!]
    \centering
    \subfigure[$\lambda = 0.5$]{\includegraphics[width = 0.3\textwidth]{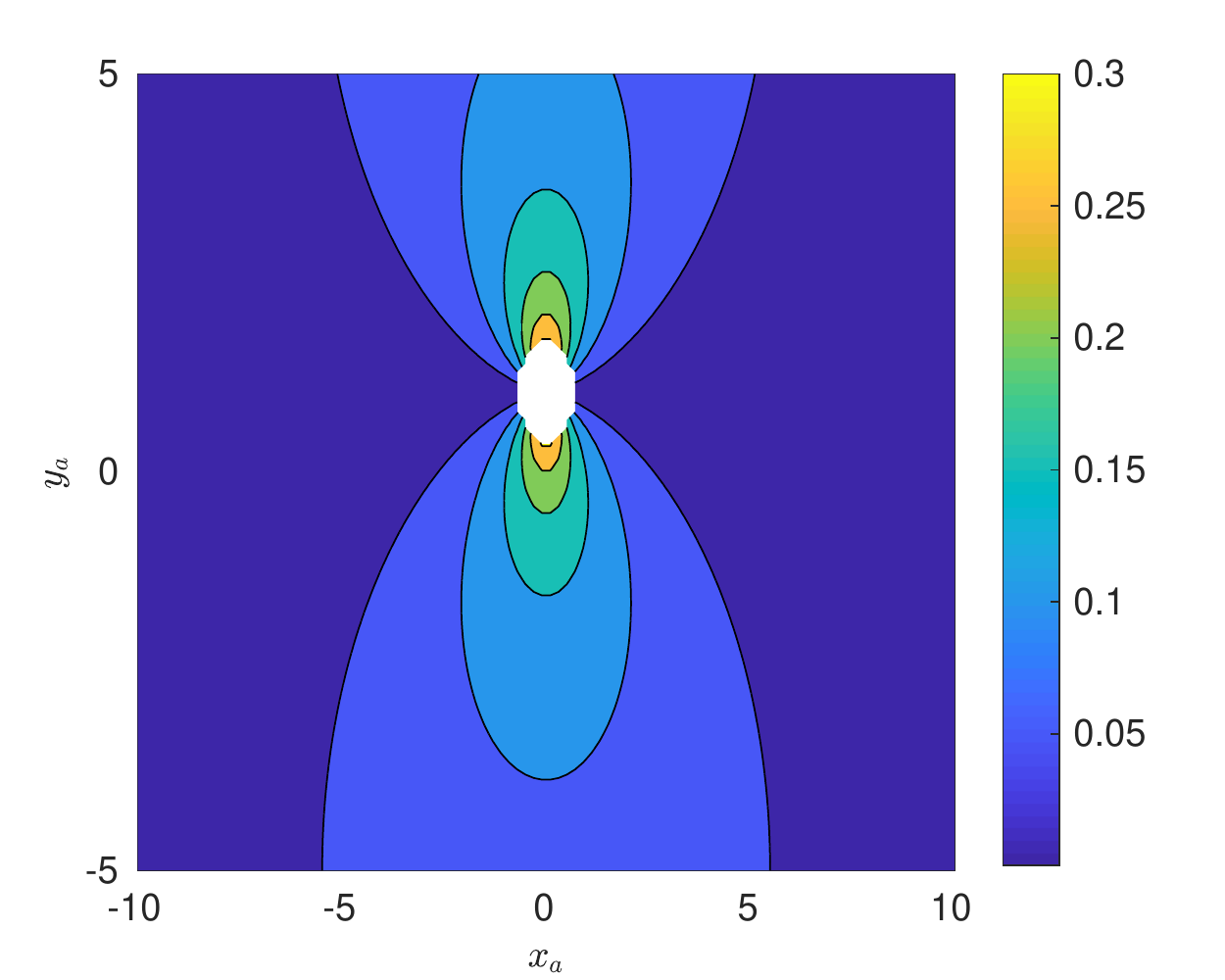}}
    \subfigure[$\lambda = 0.5$]{\includegraphics[width = 0.3\textwidth]{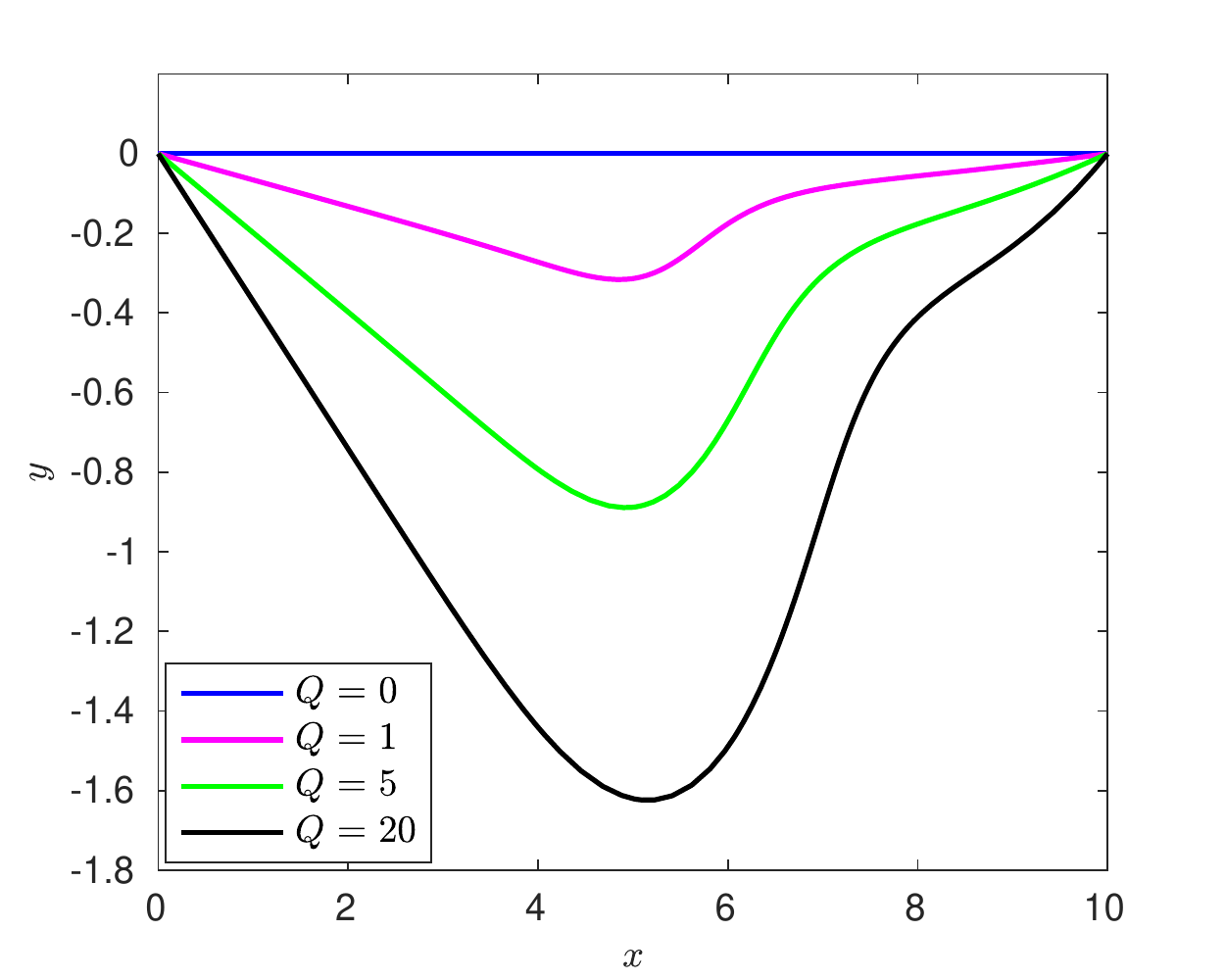}}\\
    \subfigure[$\lambda = 2$]{\includegraphics[width = 0.3\textwidth]{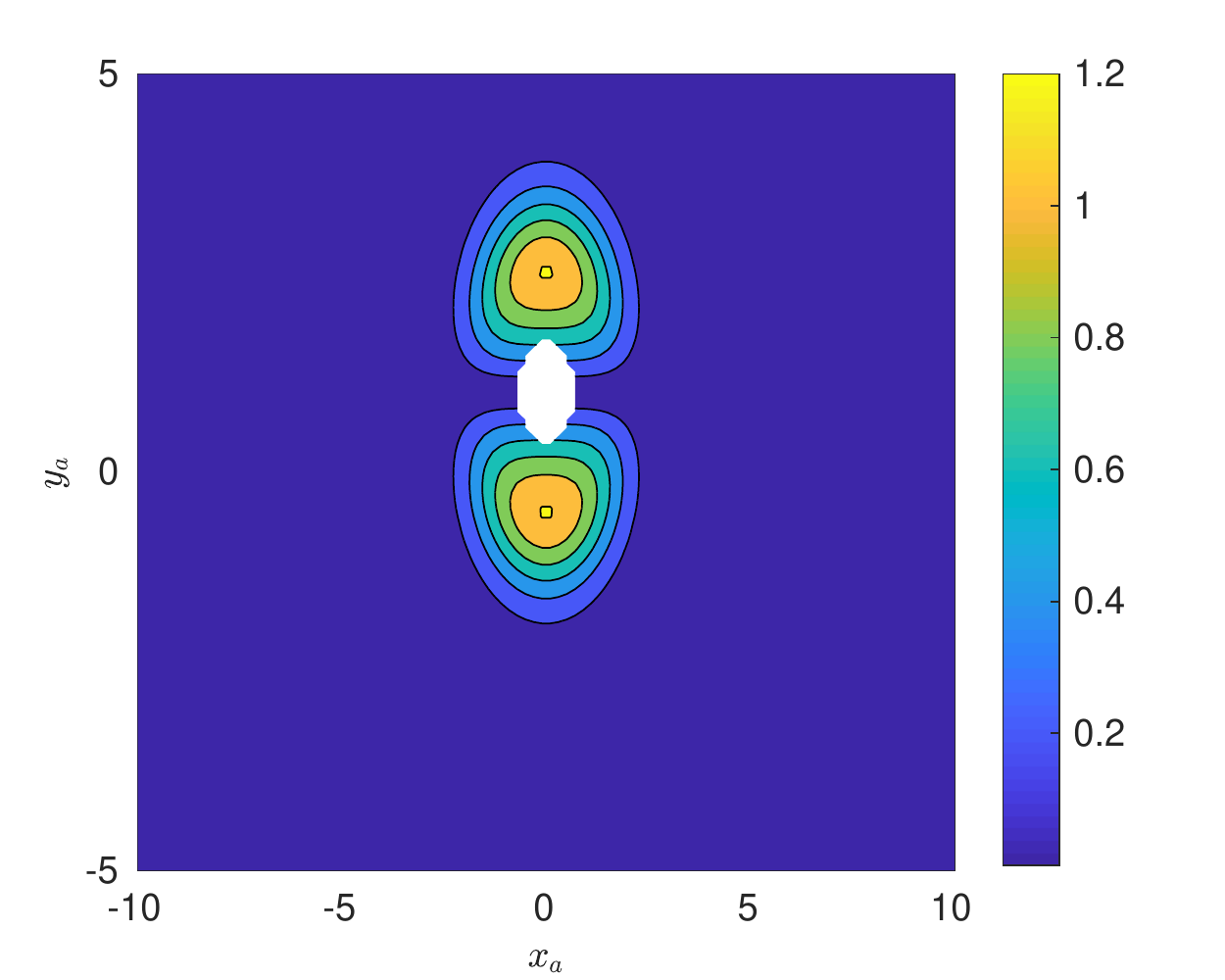}}
    \subfigure[$\lambda = 2$]{\includegraphics[width = 0.3\textwidth]{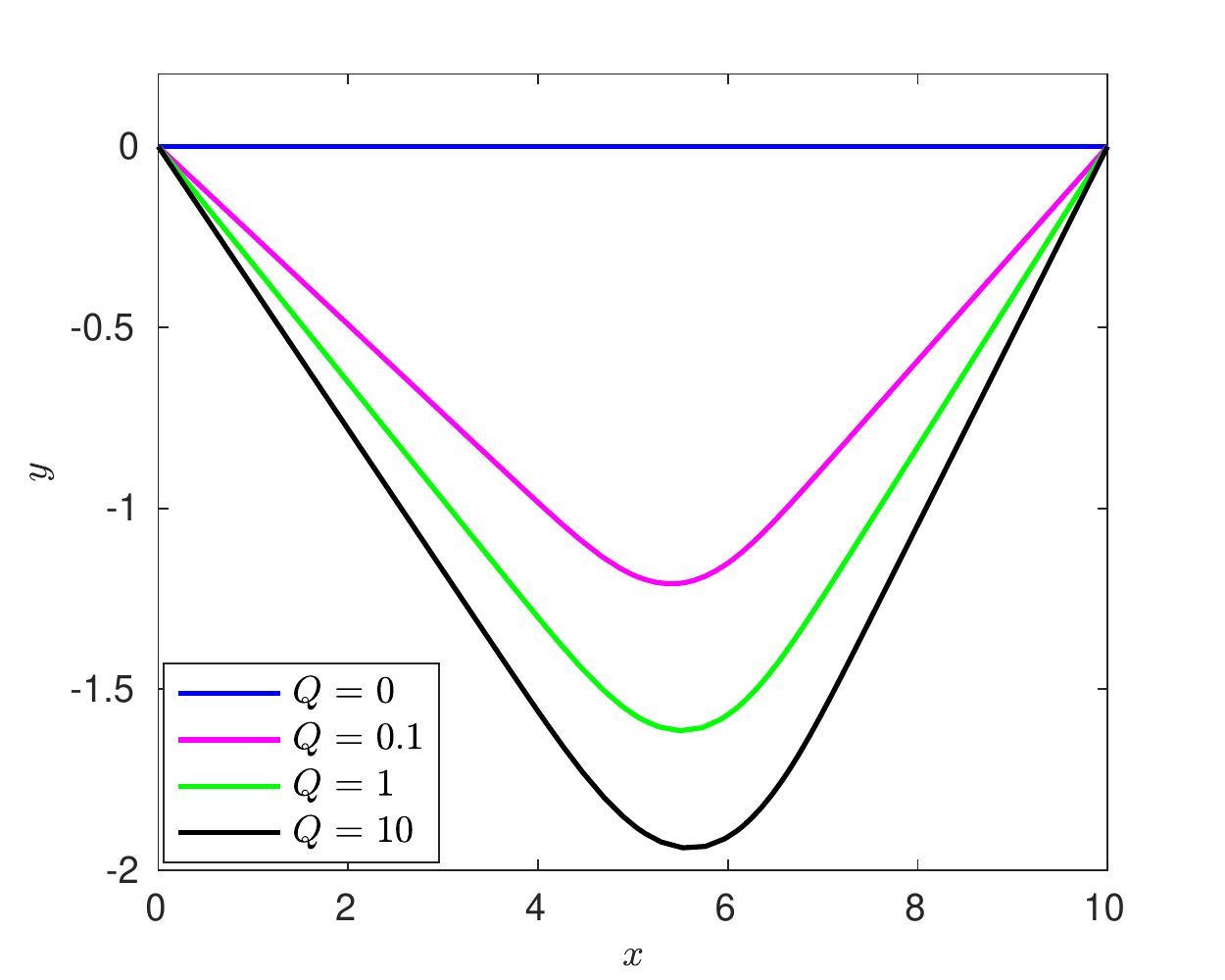}}\\
    \subfigure[$\lambda = 4$]{\includegraphics[width = 0.3\textwidth]{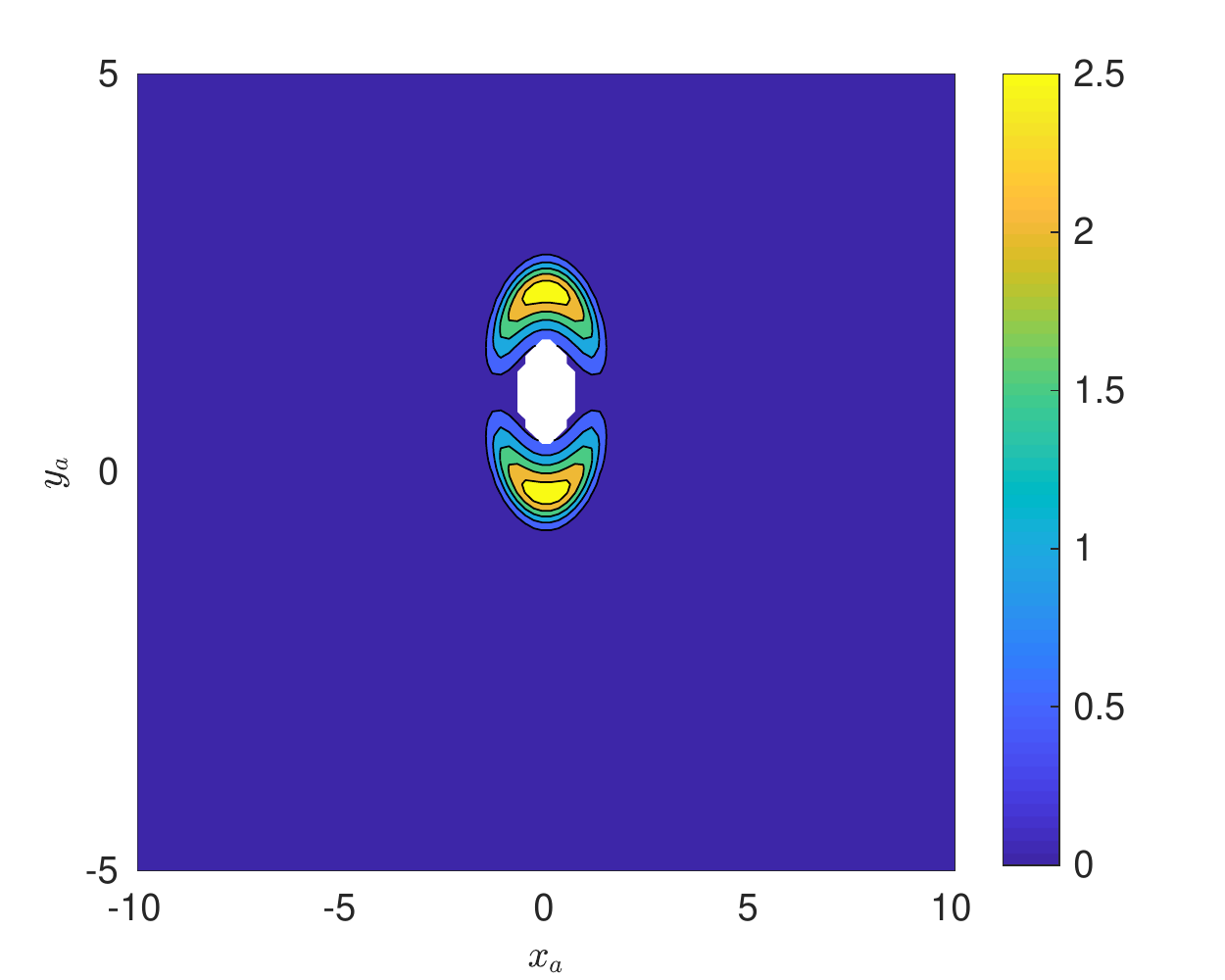}}
    \subfigure[$\lambda = 4$]{\includegraphics[width = 0.3\textwidth]{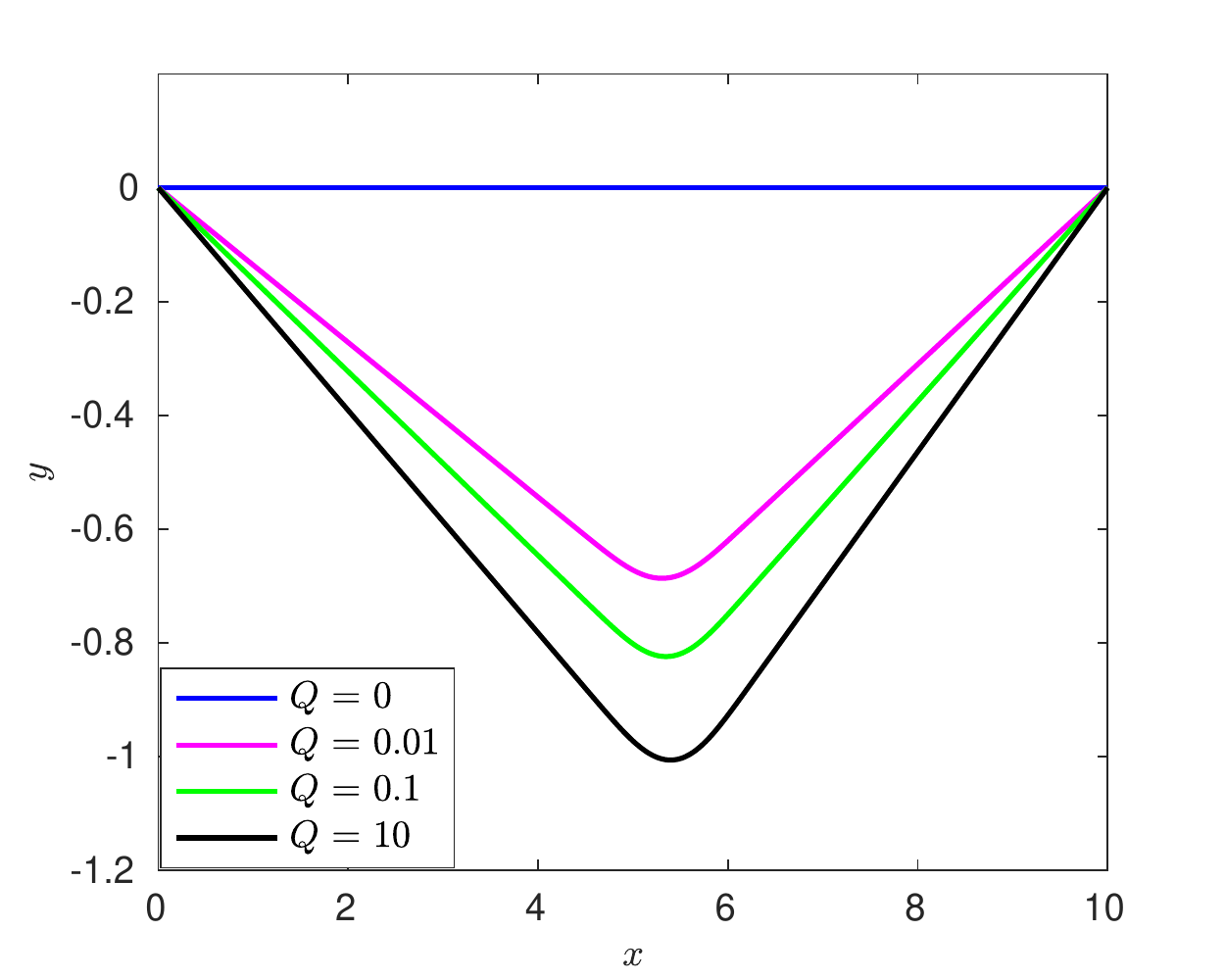}}
    \caption{Absolute values of the RCS (a, c, e) and the optimal trajectories (b, d, f) for different constraint forms.}
    \label{fig:2D_sensi_p}
\end{figure}

Figure~\ref{fig:2D_oneobs} shows the results obtained for the time-optimal path planning problem with $(a_0,b_0) = (0,0)$, $(a_f,b_f) = (0,10)$, $x_o = 5$, $c = 2$, and $v_a = 1$. 
The nominal value of the uncertain parameter $v_o$ is chosen to be 0.25.
The cost function in (\ref{eq:CDP_cost}) is minimized with $\phi(x(t_f),t_f) = t_f$, $L(x(t),u(t),t) = 0$, given the dynamics (\ref{eq:xa_dyn})-(\ref{eq:yo_dyn}), and the constraint (\ref{eq:2D_constraint}).
The optimal control package GPOPS-II \cite{Patterson2014} is used to numerically solve the optimal control problem.
From Figure~\ref{fig:2D_oneobs}(a), it can be observed that as the weighting factor $Q$ for the RCS term in the cost function increases, the optimal trajectory becomes more conservative (the distance to the obstacle is greater), and the magnitude of the RCS reduces along the trajectory (see Figure~\ref{fig:2D_oneobs}(b)).
Figure~\ref{fig:2D_oneobs}(c) suggests that the conservative trajectories have higher constraint sensitivity, and further corroborates the underlying intuition behind introducing the relevance function.

For this particular example, five different relevance functions: 1) $e^{-z^2}$ (Gaussian); 2) $\text{max}(0,1-|z|)$ (Hat function); 3) $s(z)$ (Logistic function); 4) $1/(1+z^2)$; and 5) $1/(1+|z|)^2$; were also evaluated to observe the behavior of the constraint desensitized trajectories. 
All the aforementioned functions provide conservative trajectories similar to the ones in Figure~\ref{fig:2D_oneobs}(a), with slight differences in curvature. 
The results are not presented in the interest of brevity.
Analyzing different relevance functions and their impact on constraint desensitization in general optimal control problems is a separate study meant for future work.

\subsection{Dependency on the Constraint Form}
\label{subsec:constraint_forms}

In this section, the effect of the form of the constraint function over the behavior of the optimal trajectories obtained from constraint desensitized planning is investigated. 
To this end, the constraint function in (\ref{eq:2D_constraint}) is expressed alternatively as
\begin{align}
    g_\lambda(x) = r_o^\lambda - \left[(x_a - x_o)^2 + (y_a - y_o)^2\right]^{\lambda/2} \leq 0, \label{eq:2D_constraint_lambda}
\end{align}
where $\lambda > 0$.
We first analyze the RCS plots shown in Figures~\ref{fig:2D_sensi_p}(a), \ref{fig:2D_sensi_p}(c), and \ref{fig:2D_sensi_p}(e).
The simulation parameters remain the same as the ones used for the results in Figure~\ref{fig:2D_sensi}.
The general expression for RCS, for any $\lambda > 0$, is given by
\begin{align}
    S_r = -\underbrace{\rho(g_\lambda(x))}_\text{relevance term} \times \underbrace{\lambda(y_a - y_o)t\left[(x_a - x_o)^2 + (y_a - y_o)^2\right]^{\lambda/2 - 1}}_\text{constraint sensitivity} \label{eq:gen_SL}
\end{align}
Note that, for $\lambda>0$, the relevance term (see (\ref{eq:gen_SL})) decays exponentially as the agent moves away from the obstacle. 
For $\lambda>2$, the constraint sensitivity term increases super-linearly with separation between the agent and the obstacle.
Consequently, RCS decays with Euclidean distance and becomes prominent as the agent gets closer to the obstacle.
Also, as $\lambda$ increases, the decay (logistic) term overpowers the constraint sensitivity term and the penalty region around the obstacle shrinks.

The optimal trajectories given the constraint (\ref{eq:2D_constraint_lambda}), for $\lambda=$ 0.5, 2, 4, and with different weights ($Q$) in the RCS cost can be seen in Figures \ref{fig:2D_sensi_p}(b), \ref{fig:2D_sensi_p}(d), and \ref{fig:2D_sensi_p}(f), respectively.
The simulation parameters are the same as the ones used for the results in Figure~\ref{fig:2D_oneobs}.
Similar to the results obtained for $\lambda=1$, the optimal trajectories become conservative for all values of $\lambda$, as $Q$ increases. 
However, it is noted that the behavior of these optimal trajectories vary. 
As $\lambda$ increases (for $\lambda \geq 1$), the curvature of the desensitized trajectories reduces. 
For this particular example, under turn radius constraints, the designer can alternatively tune the value of $\lambda$ to obtain the desired trajectory shape.

\subsection{The Car vs. Train Problem}

\begin{figure}
    \centering
    \includegraphics[width = 0.4\textwidth]{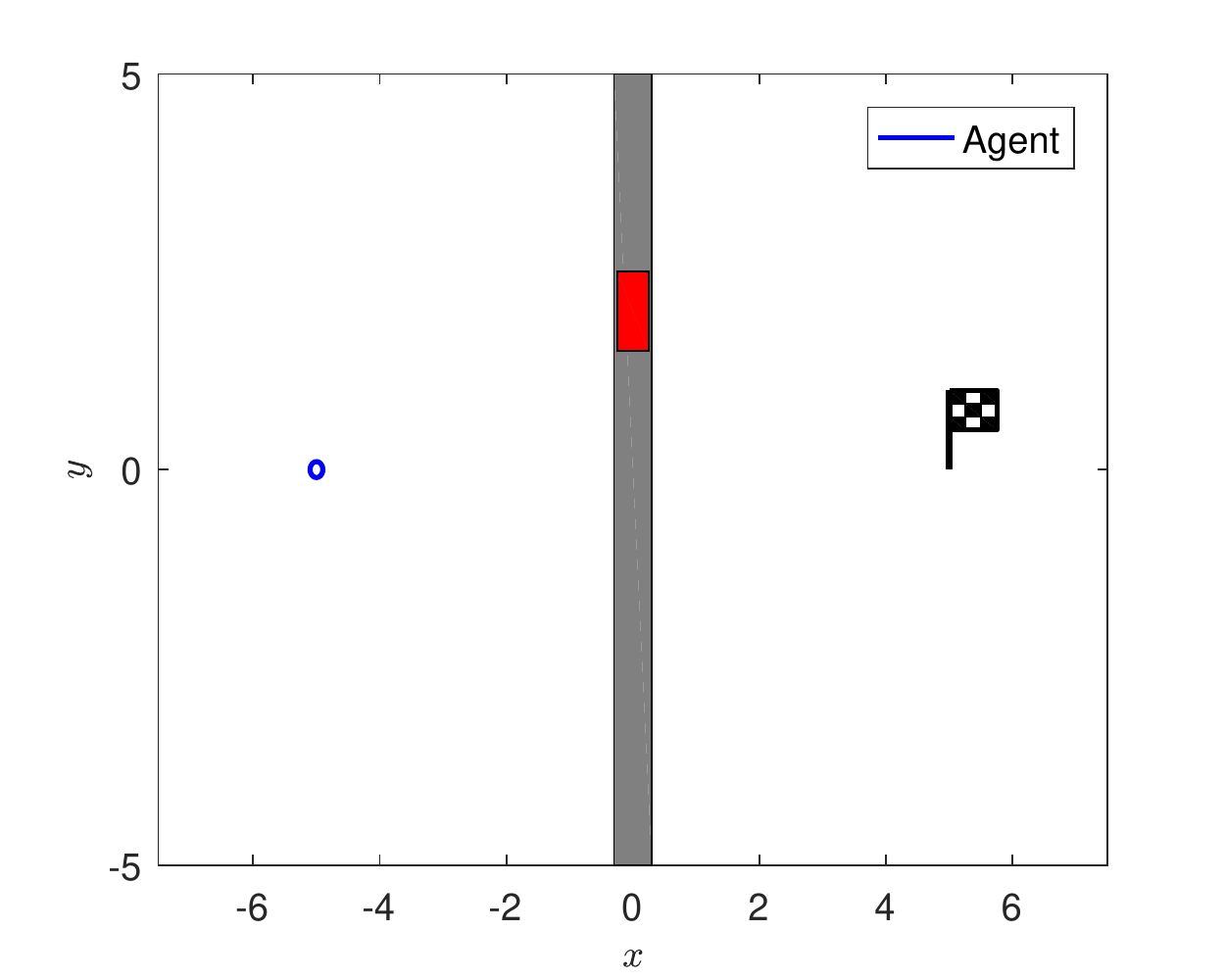}
    \caption{Schematic of the Car vs. Train problem}
    \label{fig:1D_schematic}
\end{figure}

Consider the 1D version of the problem described in Section \ref{subsec:2D_PP}, where an agent (car) is restricted to move along the x-axis, and the obstacle (train) is moving along the y-axis. 
Note that the dynamics for the obstacle remain the same, given in (\ref{eq:yo_dyn}), while the agent dynamics takes the form 
\begin{align*}
    \dot{x}_a(t) = u(t), \quad x_a(0) = a_0,
\end{align*}
where $u(t) \in [0,v^{\max}_a]$, and $y_a(t) = 0$, $t \in [0,t_f]$.
In this case, the state vector $x = [x_a,y_o]^\top$ is two-dimensional. 
A schematic of the problem setup is presented in Figure \ref{fig:1D_schematic}.

Similar to the previous experiments, we assume that the agent's primary task is to minimize travel time whilst reducing risk of collision under uncertainty in the obstacle's speed. 
It is intuitive to expect that the desensitized solution will ensure that the distance between the train and the car is sufficiently large during the event of crossing the rail track. 
Computing an RCS cost using the constraint (\ref{eq:2D_constraint}) is found to provide a desensitized solution that drives the agent to reach the target point in minimum time, regardless of the distance between the agent and the obstacle (i.e, beyond the safe distance $r_o$).

The discrepancy between the desensitized solution for the RCS, obtained using the constraint in (\ref{eq:2D_constraint}) and the intuitive solution can be understood by considering the behavior of a real-world driver.
The expression in (\ref{eq:2D_constraint}), although a valid constraint, does not capture a driver's perception of the collision constraint in this problem. 
Effectively, the train's motion along the rail track is of no consequence to the driver, except when he is crossing the track. 
It is during this crossing phase that the driver would ensure sufficient separation (at least the safe distance $r_o$) between the train and the car to prevent collision.

\begin{figure}[htb!]
    \centering
    \subfigure[A super-Gaussian approximation to a Boxcar function]{\includegraphics[width = 0.35\textwidth]{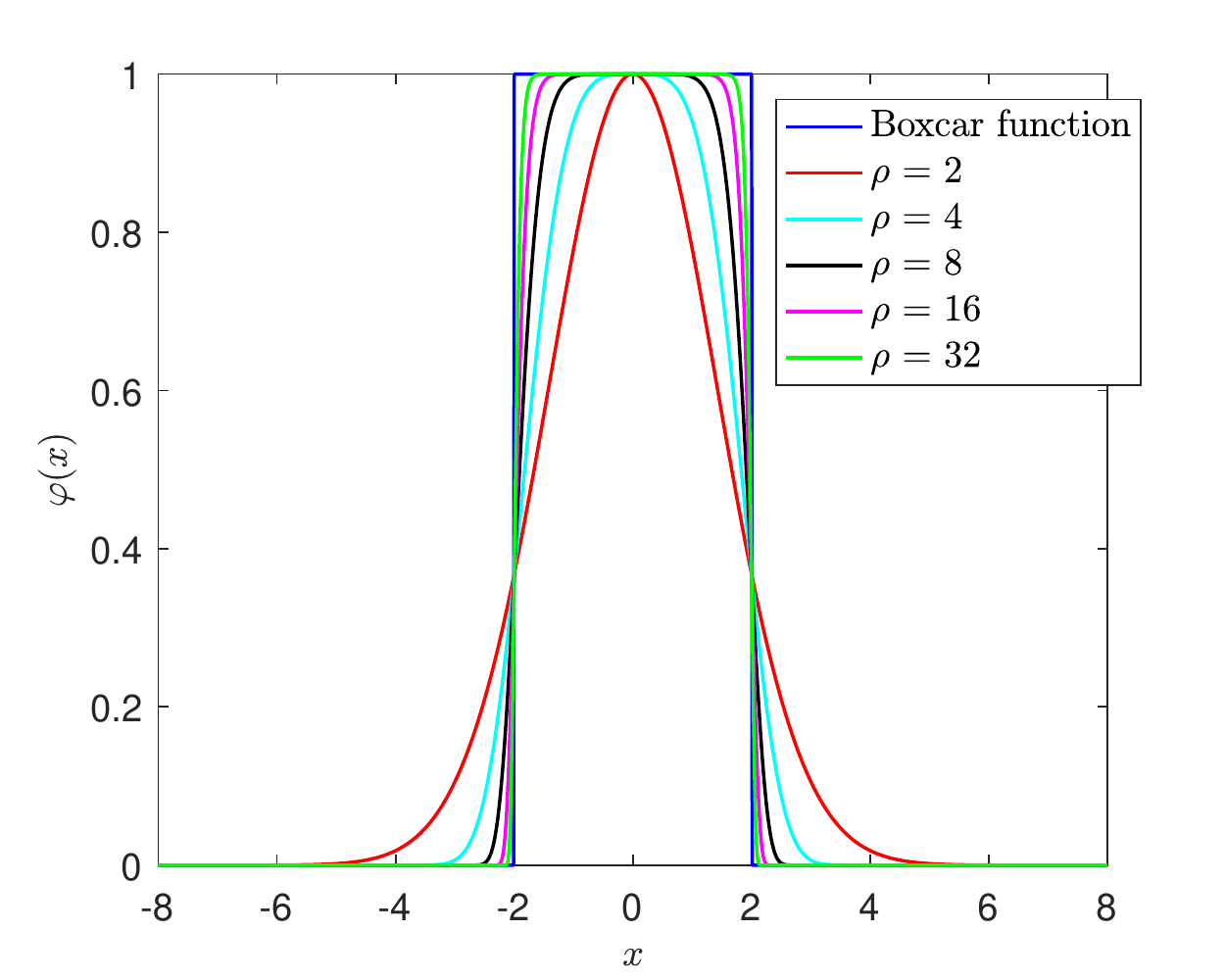}}
    \subfigure[Optimal control for different levels of constraint desensitization]{\includegraphics[width = 0.35\textwidth]{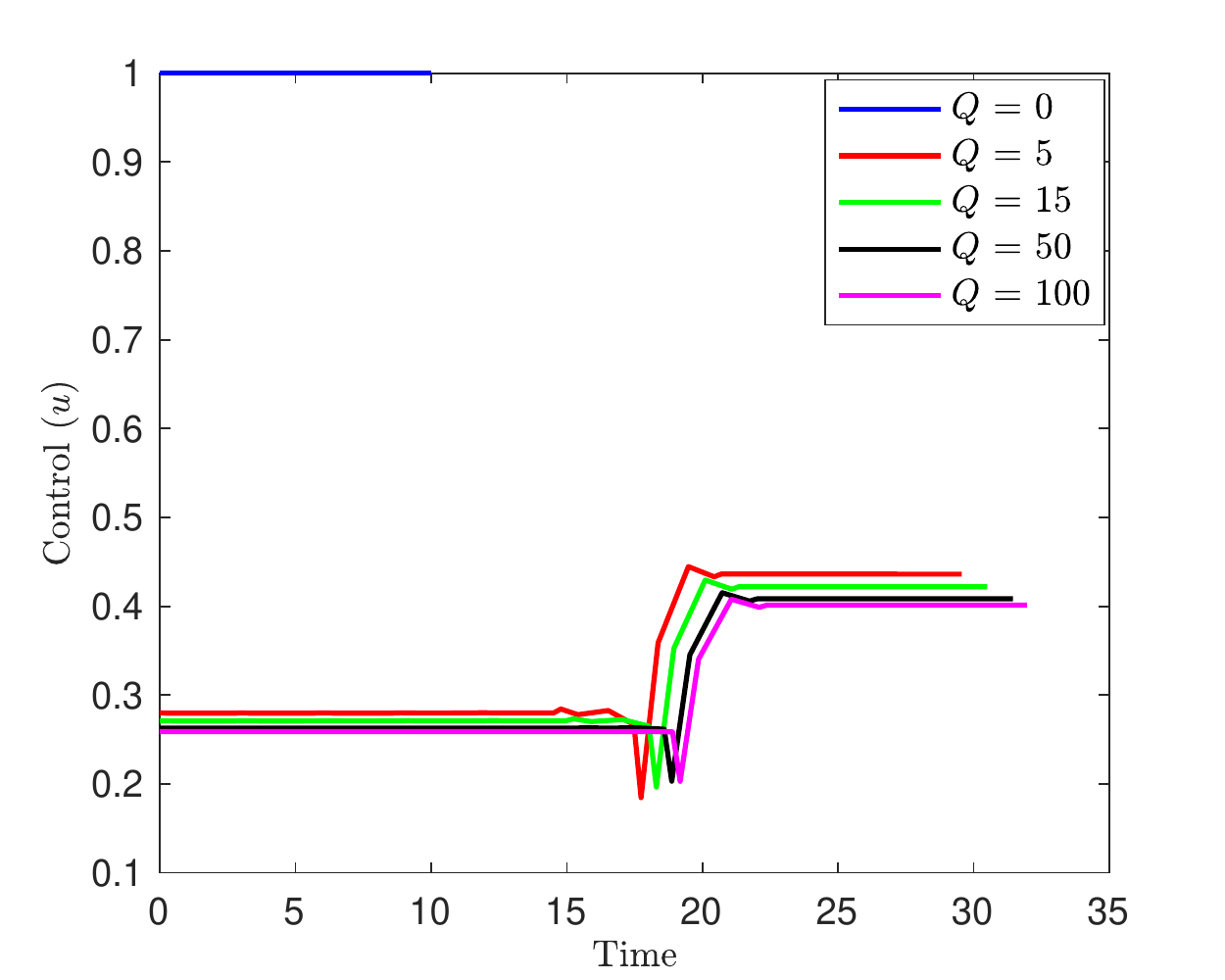}}
    \caption{The car vs. train problem}
    \label{fig:1D}
\end{figure}

This motivates us to propose a constraint of the form,
\begin{align}
    \mathds{1}[|x_a(t) - x_o| \leq w_o] \underbrace{[r_o^2 - (y_a - y_o(t))^2]}_{g_y(y_o)} \leq 0, \label{eq:1D_constraint}
\end{align}
where $\mathds{1}[\cdot]$ is the indicator function, and $2w_o$ is the width of the track.
Note that $\mathds{1}[|x_a(t) - x_o| \leq w_o]$ is a boxcar function which makes the constraint function in (\ref{eq:1D_constraint}) non-smooth.
To this end, we suggest to use the so-called \emph{super-Gaussian} \cite{beirle2017}
\begin{align}
    \varphi(x_a(t);x_o,w_o) = \exp\left(-\left[\frac{x_a(t) - x_o}{w_o}\right]^\gamma\right),
\end{align}
where $\gamma \in 2\mathbb{Z}^+$ (set of positive even numbers), as an approximation to the boxcar function.
From Figure~\ref{fig:1D}(a), it can be observed that as $\gamma \rightarrow \infty$, $\varphi$ converges to a boxcar function.
Subsequently, the RCS for (\ref{eq:1D_constraint}) with a super-Gaussian approximation can be obtained as
\begin{align}
    S_r = \rho\big(\varphi(x_a(t);x_o,w_o)g_y(y_o)\big)\frac{\partial \big(\varphi(x_a(t);x_o,w_o)g_y(y_o)\big)}{\partial v_o}. \label{eq:1D_rcs}
\end{align}
Figure~\ref{fig:1D}(b) shows the optimal control for different levels of constraint desensitization, while employing RCS in (\ref{eq:1D_rcs}) with the same simulation parameters as before, except now $\gamma = 20$, $v_a^{\max} = 1$.
It is observed that when there is no penalty on RCS, the car is dangerously close to the train at the crossing.
The result further confirms that the RCS in (\ref{eq:1D_rcs}), when penalized appropriately, allows the car to maintain a safe distance while crossing the track to avoid collision under uncertainty in the speed of the train.
This example indicates that an appropriate constraint function is crucial for the success of the proposed approach.


\subsection{Trade-off Studies with Multiple Obstacles} 
\label{sec:multi_obs}

\begin{figure}[htb!]
    \centering
    \includegraphics[width = 0.35\textwidth]{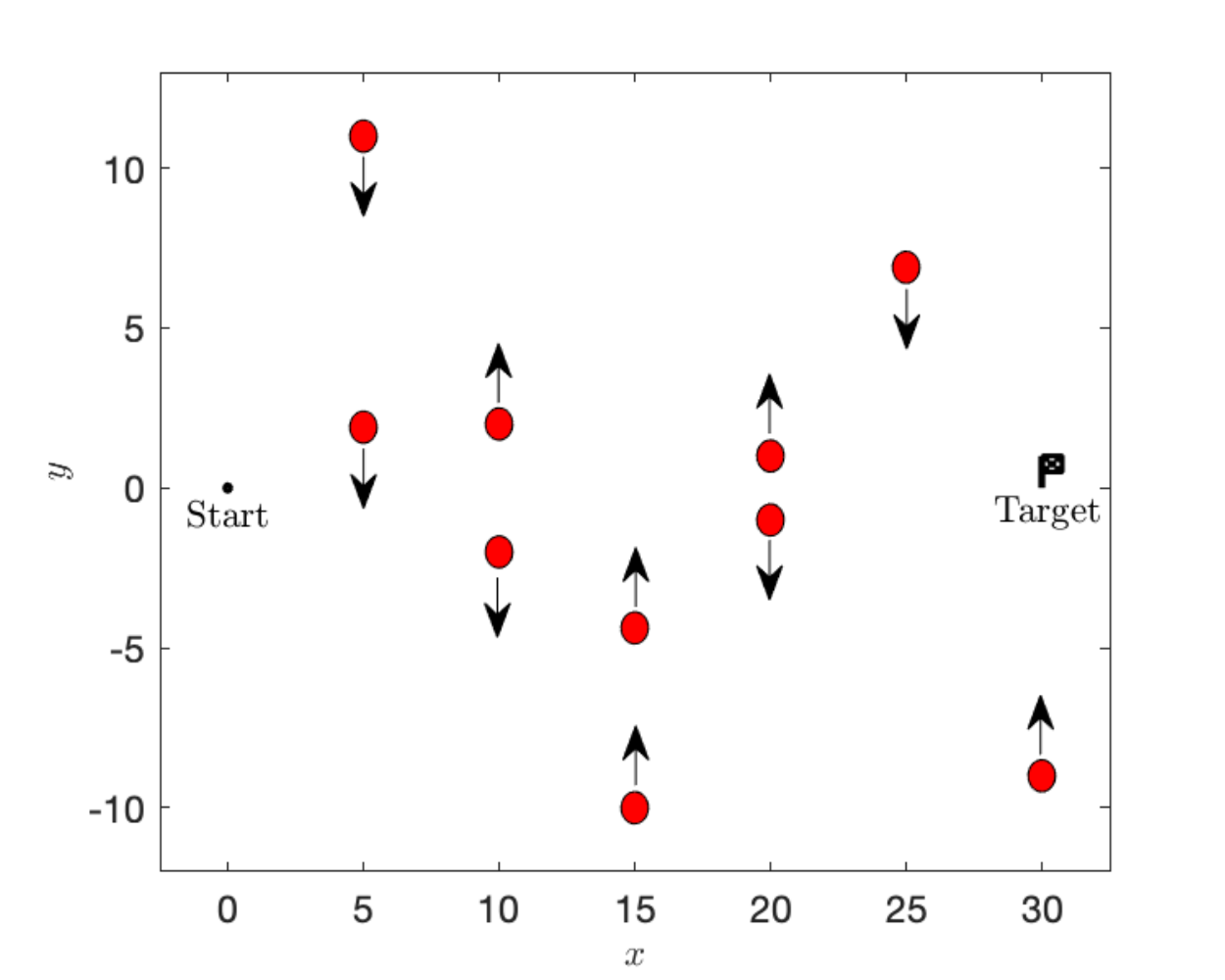}
    \caption{Schematic of an uncertain multi-obstacle environment}
    \label{fig:multiobs}
\end{figure}

In this section, the proposed approach is evaluated in instances involving multiple dynamic obstacles moving with uncertain velocities.  
The dynamics of the agent follow (\ref{eq:xa_dyn}), (\ref{eq:ya_dyn}).
In addition to the agent's heading $\theta$, its speed $v_a \in [0,1]$ is included as a control input.
Starting at location $(0,0)$, the agent is tasked with reaching the target location $(30,0)$ in minimum time while avoiding the obstacles. 
We consider four different instances with the number of obstacles $N \in \{2, 3, 5, 10\}.$
The obstacles are all assumed to be identical and their movement is restricted to be parallel to the y-axis with their speed $v_o$ being the uncertain parameter. 
The nominal value of $v_o$ is $0.25$. 
A schematic of the environment, containing the initial positions of the agent and the obstacles, and the directions of the obstacles' nominal velocity vectors for the case of $N=10$, is shown in Figure~\ref{fig:multiobs}.

\begin{figure}[htb!]
    \centering
    \subfigure[Collision probability ($P_c$)]{\includegraphics[width = 0.35\textwidth]{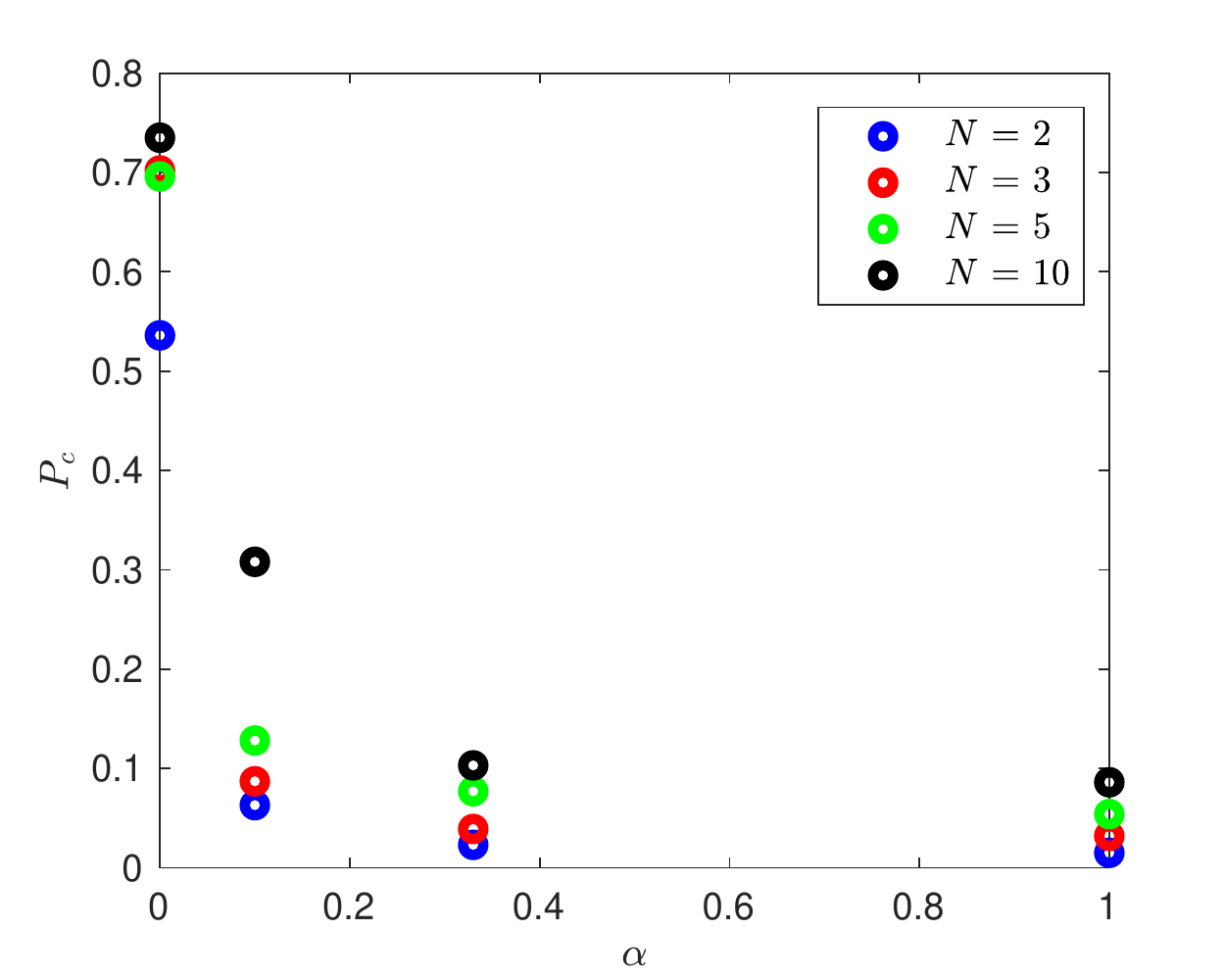}}
    \subfigure[Travel time ($t_f$)]{\includegraphics[width = 0.35\textwidth]{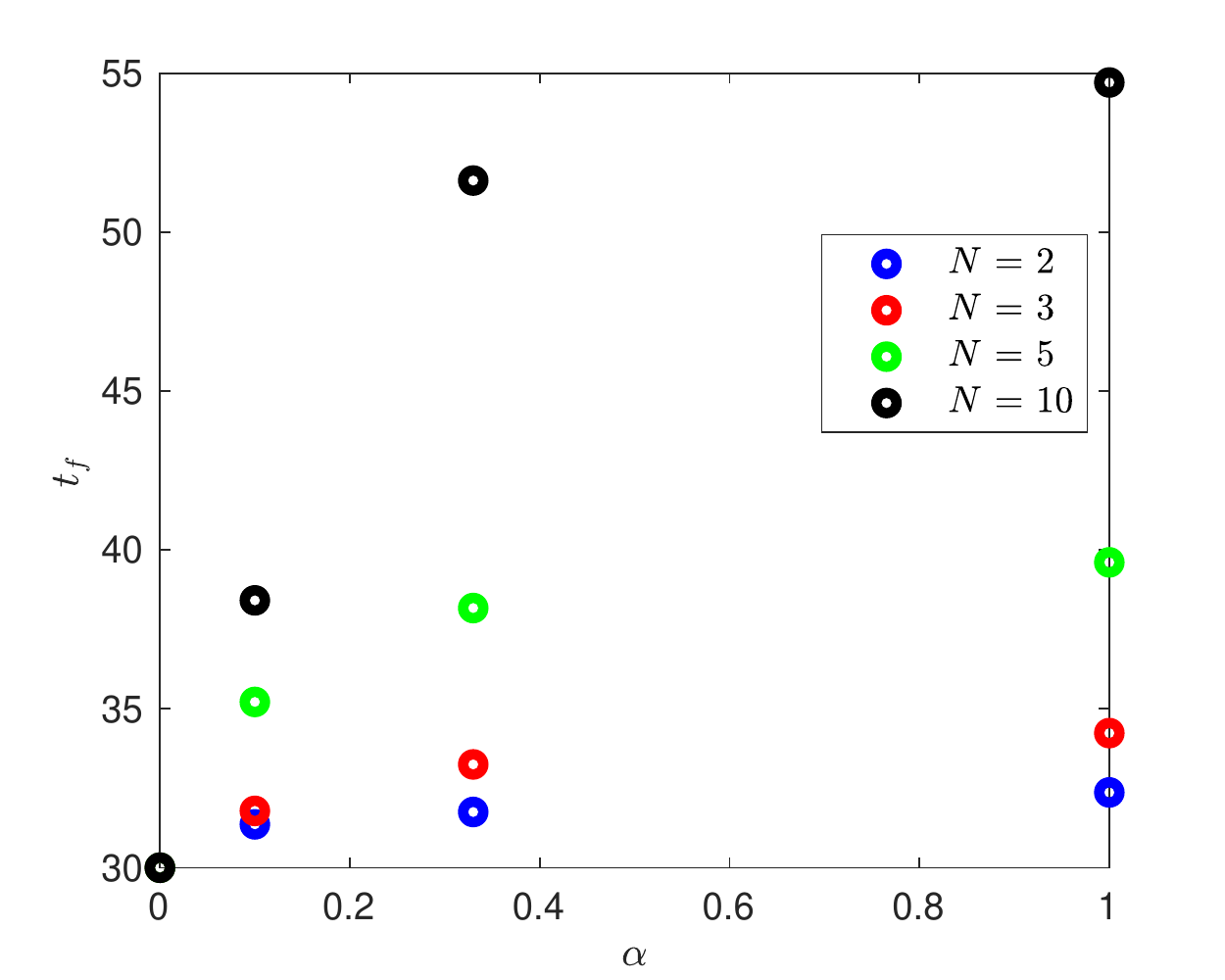}}
    \caption{Trade-offs plots for instances with different number of obstacles}
    \label{fig:multiobs_prob_tf}
\end{figure}

Note that at each instance, there are $N$ constraints enforcing collision avoidance, and the speed of the associated obstacle is the uncertain parameter.
Ignoring the zeros in the matrix $S_r(t)$, all the sensitivity terms are weighted equally by choosing the $Q$ in the RCS cost in (\ref{eq:CDP_cost}) to be of form $Q = \alpha I_N$.
Each instance involves four levels of penalization of the RCS cost with $\alpha =$ 0 (blue, no penalty), 0.1 (magenta), 0.33 (green), 1 (black). 
From Figure~\ref{fig:multiobs_prob_tf}(b), and for different instances ($N = 2,3,5,10$), it can be observed that as $\alpha$ increases, the agent takes longer paths, essentially trying to avoid obstacles while maintaining some safety buffer.
To characterize safety, collision probabilities were computed by running Monte Carlo simulations on the optimal trajectory, obtained from GPOPS-II, while propagating the dynamics in an open-loop fashion for 1000 samples. 
In the Monte Carlo simulations the variation in each of the obstacle's speed $\Delta v_o$ is obtained by sampling from a normal distribution $N(0,\sigma^2)$ with $\sigma^2 = 0.1$.
The trade-off between travel times ($t_f$) and collision probabilities ($P_c$) for the four instances are shown in Figure~\ref{fig:multiobs_prob_tf}.
From Figure~\ref{fig:multiobs_prob_tf}, it is seen that penalizing RCS (for higher $\alpha$ values) yields safer trajectories, which mitigate the chance of constraint violation in uncertain environments while trading off optimality (travel time).
For instances with $N=2,3$, a 95\% reduction in collision probability is achieved for a 10\% trade-off in travel time.
Due to the cumulative effect of increasing the number of obstacles obstacles and the number of uncertain parameters on the RCS cost (that measures the risk of constraint violation), the travel times for safer trajectories are seen to increase with $N$.

It is observed that the computation times for the tested instances are of the same order of magnitude (a few milliseconds).
The approach is limited by the efficiency of the chosen optimal control solver.
The regularizer has no guarantees in terms of convexity, and consequently the optimizer may converge to a local minimum. 
Depending on the initialization, the homotopy class of the obtained trajectories may vary.
For the above simulations, we report the optimal trajectory among the ones obtained from different initializations. 
While in the above simulations the obstacles are restricted to follow simple paths parallel to the y axis, it is important to note that the regularizer can be derived for arbitrary obstacle motion as long as its dynamics are known and the uncertain parameters are identified.


\section{Conclusion} 
\label{sec:conclusion}

A sensitivity function-based regularizer is introduced to obtain conservative solutions that avoid constraint violation under parametric uncertainties in optimal control problems.
Using the fact that collision avoidance can be expressed as a state constraint, the approach is applied for path planning problems involving dynamic uncertain obstacles.
The proposed regularizer is first analyzed on simple problems to study its characteristics and to identify its limitations.
It is observed that the form of the constraint function used to construct the regularizer affects the behavior of the trajectories.
The results on environments with as many as ten dynamic obstacles indicate that safety can be enhanced with an acceptable trade-off in optimality.


\section*{Acknowledgments}

This work has been supported by NSF award CMMI-1662542.


\end{document}